\documentclass[10pt,twocolumn,letterpaper]{article}

\usepackage[pagenumbers]{cvpr} 

\usepackage{graphicx}
\usepackage{amsmath}
\usepackage{amssymb}
\usepackage{booktabs}
\usepackage{cuted}
\usepackage[table,xcdraw,dvipsnames]{xcolor}
\usepackage{textcomp}
\usepackage{tabularx}
\newcolumntype{Y}{>{\centering\arraybackslash}X}
\usepackage{microtype}
\usepackage{wrapfig}
\usepackage{MnSymbol,bbding,pifont}
\usepackage{graphbox}
\usepackage{soul}

\newcommand{\bc}{\mathbf{c}}

\newcommand{\bP}{\mathbf{P}}

\newcommand{\bw}{\mathbf{w}}

\newcommand{\bz}{\mathbf{z}}

\newcommand{\bepsilon}{\boldsymbol{\epsilon}}

\newcommand{\figref}[1]{Fig.~\ref{#1}}
\newcommand{\secref}[1]{Section~\ref{#1}}

\newcommand{\tabref}[1]{Table~\ref{#1}}

\usepackage{amsfonts}

\makeatletter
\DeclareRobustCommand\onedot{\futurelet\@let@token\@onedot}
\def\@onedot{\ifx\@let@token.\else.\null\fi\xspace}
 
\def\ie{i.e\onedot}

\def\wrt{wrt\onedot}

\def\etal{et~al\onedot}

\makeatother

\usepackage{xcolor}
\definecolor{darkred}{rgb}{0.7,0.2,0.1}
\definecolor{darkgreen}{rgb}{0,0.7,0}
\definecolor{orange}{RGB}{255,127,0}
\definecolor{ourpurple}{RGB}{127,127,204}
\definecolor{palgreen}{RGB}{51,179,179}
\definecolor{magenta}{RGB}{199,21,133}

\graphicspath{{figures/}}
\usepackage[dvipsnames]{xcolor}

\definecolor{lightblue}{RGB}{171,230,237}
\definecolor{lightgreen}{RGB}{193,247,193}
\definecolor{tabfirst}{rgb}{1, 0.7, 0.7} 
\definecolor{tabsecond}{rgb}{1, 0.85, 0.7} 
\definecolor{tabthird}{rgb}{1, 1, 0.7} 

\makeatletter
\renewcommand\paragraph{\@startsection{paragraph}{4}{\z@}%
	{0.75ex \@plus.5ex \@minus.2ex}%
	{-1em}%
	{\normalfont\normalsize\bfseries\maybe@addperiod}}
\newcommand{\maybe@addperiod}[1]{#1\@addpunct{.}}
\makeatother

\definecolor{cvprblue}{rgb}{0.21,0.49,0.74}
\definecolor{cvprurl}{rgb}{0.98,0.27,0.63}
\definecolor{teaserpurple}{rgb}{0.6,0.63,0.89}
\definecolor{teaseryellow}{rgb}{0.94,0.93,0.70}
\definecolor{teaserblue}{rgb}{0.41,0.59,0.90}
\usepackage[pagebackref,breaklinks,colorlinks,citecolor=cvprblue,urlcolor=cvprurl]{hyperref}

\newcommand\rurl[1]{%
  \href{http://#1}{\nolinkurl{#1}}%
}

\def\paperID{6583} 
\def\confName{CVPR}
\def\confYear{2024}

\title{CAD\,\raisebox{-0.4ex}{\includegraphics[scale=0.03]{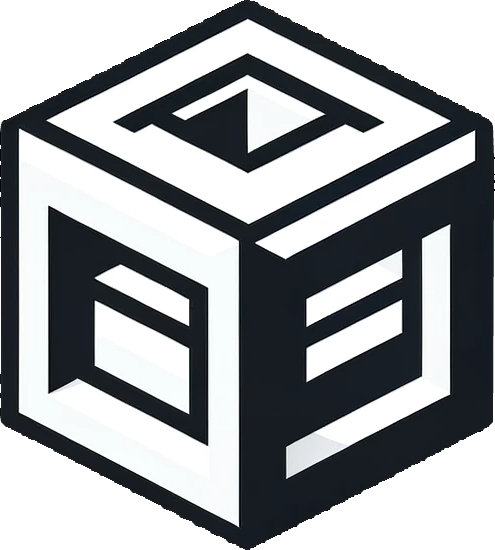}}\,: Photorealistic 3D Generation via Adversarial Distillation}

\author{Ziyu Wan$^{1,2}$ \quad 
Despoina Paschalidou$^{2}$ \quad  Ian Huang$^2$  \quad Hongyu Liu$^{3}$ \quad Bokui Shen$^{2}$  \\  Xiaoyu Xiang \quad Jing Liao$^{1*}$ \quad  Leonidas Guibas$^{2}$   \\
	$^1$City University of Hong Kong \quad    
	$^2$Stanford University \quad    
        $^3$HKUST  \quad   \quad 
	\\
        \textbf{\rurl{raywzy.com/CAD}}
	}

\begin{document}

\twocolumn[{
\renewcommand\twocolumn[1][]{#1}
\maketitle
    \vspace{-3.0em}
    \setlength\tabcolsep{0.5pt}
    \centering
    \includegraphics[width=1.0\textwidth]{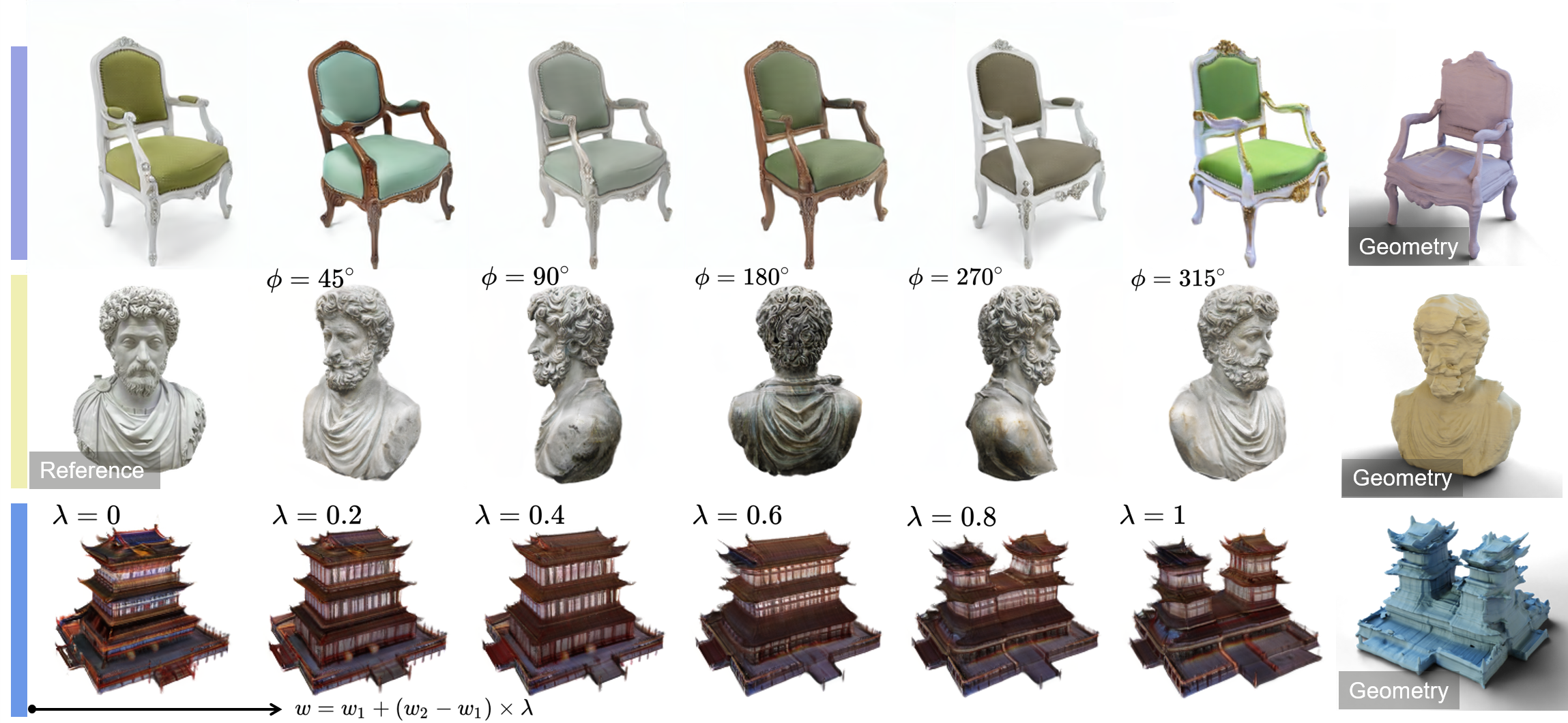}
    \vspace{-1.5em}
    \captionsetup{type=figure} 
    \captionof{figure}{ CAD leverages pretrained diffusion models to generate photorealisitc 3D contents based on a single input image and the text prompt, enabling different applications including \textcolor{teaserpurple}{$\bullet$} diversified generation, \textcolor{teaseryellow}{$\bullet$} single-view reconstruction by inversion and \textcolor{teaserblue}{$\bullet$} 3D interpolation.}
    \vspace{0.5em}
    \label{figure:teaser}
}]

\maketitle
\begin{NoHyper}\footnotetext{\hspace{-1.25em}* Corresponding author.}\end{NoHyper}
\begin{abstract}
The increased demand for 3D data in AR/VR, robotics and gaming applications, gave rise to powerful generative pipelines capable of synthesizing high-quality 3D objects. Most of these models rely on the Score Distillation Sampling (SDS) algorithm to optimize a 3D representation such that
the rendered image maintains a high likelihood as evaluated by a pre-trained diffusion model. 
However, finding a correct mode in the high-dimensional distribution produced by the diffusion model is challenging and often leads to issues such as over-saturation, over-smoothing, and Janus-like artifacts. In this paper, we propose a novel learning paradigm for 3D synthesis that utilizes pre-trained diffusion models. Instead of focusing on mode-seeking, our method directly models the distribution discrepancy between multi-view renderings and diffusion priors in an adversarial manner, which unlocks the generation of high-fidelity and photorealistic 3D content, conditioned on a single image and prompt. Moreover, by harnessing the latent space of GANs and expressive diffusion model priors, our method facilitates a wide variety of 3D applications including single-view reconstruction, high diversity generation and continuous 3D interpolation in the open domain. The experiments demonstrate the superiority of our pipeline compared to previous works in terms of generation quality and diversity.
\end{abstract}    
\vspace{-2em}
\section{Introduction}
\label{sec:intro}



In recent years, we have witnessed an unprecedented explosion in generative models that can synthesize intelligible text \cite{Brown2020NeurIPS, Touvron2023ARXIV, OpenATI2023ARXIV}, photorealistic images \cite{Karras2020CVPR, Ramesh2021ICML, Rombach2022CVPR, wan2020bringing, Ramesh2022ARXIV, Rombach2022CVPR, Nichol2022ICML, Sauer2023ICML, Meng2023CVPR}, video sequences \cite{Skorokhodov2022CVPR, Bahmani2022ARXIV, Singer2022ARXIV, wan2022bringing}, music \cite{Dhariwal2020ARXIV, Borsos2022ARXIV, Agostinelli2023ARXIV} and 3D data \cite{chan2022efficient, Gao2022NEURIPS, Nichol2022ARXIV, poole2022dreamfusion, Jun2023ARXIV, zhang2022fdnerf, Lin2023CVPR, Zhang2023ICCV, Chen2023ICCV, zhang2023text2nerf}. In particular, when dealing with 3D data, manually creating them is a laborious endeavor that necessitates technical skills from highly experienced designers. Therefore, having systems capable of automatically generating realistic and diverse 3D contents could significantly facilitate the workflow of artists and product designers and could enable new levels of creativity through ``generative art'' \cite{Bailey2020ArtInAmerica}.


Recently diffusion models~\cite{Dickstein2015ICML, ho2020denoising, Song2021ICLR} have emerged as a powerful class of tools that can produce photorealistic images conditioned on versatile user inputs~\cite{Rombach2022CVPR, Saharia2022NEURIPS, Ramesh2022ARXIV, Zhang2023ICCV} such as text, depth maps, semantic masks etc. 
However, naively adopting them on 3D synthesis tasks is not trivial, due to the lack of large-scale, richly annotated 3D datasets. Despite the recently introduced larger
3D datasets \cite{Deitke2023ARXIV, Wu2023CVPR}, they remain significantly smaller compared to contemporary image-text datasets that typically contain billions of examples. Therefore,
a large body of works \cite{poole2022dreamfusion, Lin2023CVPR, Chen2023ICCV, Wang2023NEURIPS} explored using pre-trained 2D text-to-image diffusion models~\cite{Rombach2022CVPR, Saharia2022NEURIPS} to generate 3D data. Among the first works was
DreamFusion~\cite{poole2022dreamfusion} that introduced the Score Distillation Sampling (SDS) algorithm for learning a 3D representation, such that the rendered image from any view looks similar (\ie has high likelihood) to a sample from the pre-trained 2D diffusion model, given a text description.

Despite their impressive performance \cite{Liao2023THREEDV, Lin2023CVPR, Metzer2023CVPR, Wang2023CVPR, Chen2023ICCV}, 
SDS-based arts frequently exhibit issues such as over-saturation, over-smoothness, non-photorealism, and limited diversity, primarily attributed to the quality degradation of mode-seeking behaviors in deep generative models~\cite{Nalisnick2019ICLR}.
To alleviate these drawbacks, in a concurrent work, Wang \etal introduced Variational Score Distillation (VSD) \cite{Wang2023NEURIPS}, which is a generalized version of SDS that aims to optimize a 3D distribution to approximate the distribution defined by the diffusion model. While VSD~\cite{Wang2023NEURIPS} addressed some issues of SDS, its rendering quality still suffers from highly saturated colors hence resulting in less photorealistic generations.

In this paper, we propose Consistent Adversarial Distillation (CAD), a new approach for generating 3D objects conditioned on a text prompt and a single image, to overcome the mentioned issues. Instead of optimizing a single NeRF through score distillation, our key idea is to train a 3D generator that directly models the conditional distribution of a pre-trained diffusion model, through adversarial learning.
Although modeling the distribution of generic concepts with Generative Adversarial Networks (GANs) \cite{Goodfellow2014NIPS} may be challenging, utilizing input conditions such as text or images could effectively constrain the data distribution, since the 3D objects following specific conditions should share similar scale, shape and appearance in the canonical space, which GANs could handle pretty well \cite{Chen2016NIPS, Henzler2019ICCV, Schwarz2020NEURIPS, Niemeyer2021CVPR, chan2022efficient}. Moreover, as the generator learns a mapping from the latent to the continuous 3D distribution, our model becomes applicable to various downstream tasks, including diversified sampling, single-view reconstruction and 3D interpolation (see \figref{figure:teaser}).



However, distilling prior knowledge from a pre-trained diffusion model into a 3D GAN is not trivial. Existing attempts typically rely on large amount of high-quality and aligned data \cite{Wu2016NIPS, Schwarz2020NEURIPS, Niemeyer2021CVPR, Chan2021CVPR, Skorokhodov2022NEURIPS, Shi2023CVPR} with evenly-distributed poses \cite{Schwarz2020NEURIPS, Chan2021CVPR, Xu2022CVPR, chan2022efficient, Gu2022ICLR, SkorokhodovICLR2023}. When sampling novel images from a pre-trained 2D diffusion model, there is no guarantee that the data distribution will adequately cover all azimuth angles that fully define the shape's geometry. On the contrary, due to the existing inductive bias of the diffusion model, it is more likely to produce frontal-facing data, even after using prompt engineering or negative prompts. We resolve this by leveraging the view-dependent diffusion model of \cite{liu2023zero} and further propose several distribution pruning and refinement strategies that ensure stable training as well as diverse and high-quality samples. Finally, we employ a 3D-aware GAN~\cite{chan2022efficient} to learn a 3D generator that models the conditional distribution of a pre-trained diffusion model.
We evaluate our model on several datasets and showcase that it can generate high-quality, diverse and photorealistic 3D objects conditioned on a single image and a text prompt.

\section{Related Work}
\label{sec:related}


\paragraph{3D Generative Models}%
Over the years, several works explored combining generative models \cite{Goodfellow2014NIPS,Kingma2014ICLR, Rezende2015ICML,Dickstein2015ICML, ho2020denoising} with different 3D representations such as voxel grids \cite{Wu2016NIPS, Gadelha2017THREEDV, Henzler2019ICCV, Lunz2020ARXIV, Ibing2021ARXIV}, meshes \cite{Gao2019SIGGRAPHASIA, Gao2021SIGGRAPH, Zhang2021ICLR, Nash2020ICML}, point clouds \cite{Achlioptas2018ICML, Yang2019ICCV, Ramasinghe2020IROS, Zeng2022NEURIPS} or neural implicits \cite{Chen2019CVPR, Koo2023ICCV, Zhang2022NEURIPS, Zhang2023SIGGRAPH}. Although most of these pipelines can generate plausible 3D geometries, they require explicit 3D supervision.
To this end, many works investigated learning the 3D scene geometry and appearance through volumetric \cite{Niemeyer2020CVPR, Yariv2020NEURIPS, Mildenhall2020ECCV} and differentiable rendering \cite{wan2023learning, Liao2020CVPRa, Zhang2021ICLR}. The key advantage of these pipelines is that they can recover 3D information only from images. In this work, we introduce a model capable of generating high-quality 3D objects conditioned on a single image and a text prompt, hence effectively alleviating the need for both 3D data and multi-view images during training. 

\begin{figure*}
    \centering
    \includegraphics[width=\linewidth]{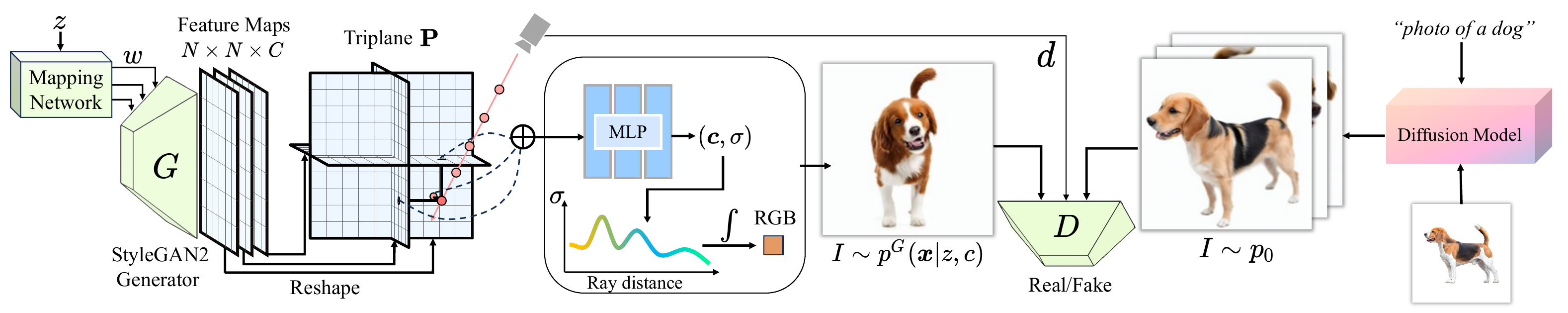}
        \vspace{-2em}
    \caption{Our adversarial distillation framework mainly comprises three parts: a StyleGAN2-based generator for approximating the target distribution, a pre-trained diffusion model for providing 2D priors based on the given input image and text prompt, and a discriminator for minimizing the distribution gap between $p^G(\boldsymbol{x} \mid \bz, \bc)$ and $ p_0\left(\boldsymbol{x}_0 \mid y\right)$. For brevity, we omit some details of the triplane generator training. Our method could effectively overcome the issues of score distillation and achieve highly photorealistic and diverse 3D generation. }
    \label{fig:architecture}
    \vspace{-1.2em}
\end{figure*}

\paragraph{3D-Aware Generative Models}%
GANs~\cite{Goodfellow2014NIPS} have demonstrated impressive capabilities on several image synthesis
\cite{Brock2019ICLR, Choi2018CVPR, Karras2019CVPR,
Karras2020CVPR} and editing \cite{Shen2020CVPR, Alharbi2020NIPS,
Ling2021NIPS, Wang2021ICLR, Isola2017CVPR, Choi2018CVPR, Brooks2023CVPR} tasks.
However, adopting them to 3D data is non-trivial as they ignore the
physics of the image formation process, hence failing to
produce 3D consistent renderings. To address this, several works \cite{Henderson2019IJCV, Henderson2020NIPS, Nguyen-Phuoc2019ICCV, Nguyen-Phuoc2020ARXIV, Gu2022ICLR, DeVries2021ICCV, Hao2021ICCV, Meng2021ICCV, Niemeyer2021THREEDV,
Zhou2021ARXIV} explored incorporating explicit 3D representations or combining GANs~\cite{Schwarz2020NEURIPS, Chan2021CVPR, chan2022efficient, Skorokhodov2022NEURIPS} with Neural Radiance Fields (NeRFs)~\cite{Mildenhall2020ECCV}.  Due to their compelling results, various follow-up works further improved various aspects of the synthesis process such as the rendering quality \cite{chan2022efficient, Gu2022ICLR, Schwarz2022NEURIPS, Xu2022CVPR, Skorokhodov2022NEURIPS} , the underlying geometry \cite{ Pan2021NEURIPS, Shi2022NEURIPS}, the editing capabilities \cite{Niemeyer2021CVPR, Yuan2022CVPR, Kwak2022ECCV, Liu2021ICCV, Tertikas2023CVPR, Haque2023ICCV, Kim2023ICCV}. 
Our generator architecture is similar to EG3D~\cite{chan2022efficient}, however our training pipeline that leverages 2D diffusion priors is novel and enables the generation of objects from arbitrary categories based on textual descriptions and image guidance.

\paragraph{3D Generation Guided by 2D Prior Models}%
Our work falls into the category of methods that leverage priors from text-to-image diffusion models \cite{Nichol2022ICML, Ramesh2022ARXIV, Rombach2022CVPR, Saharia2022NEURIPS, Aladdin2023ARXIV} for generating 3D data. DreamFusion~\cite{poole2022dreamfusion} was among the first that proposed to distill a pre-trained diffusion model ~\cite{Saharia2022NEURIPS} into a NeRF~\cite{Mildenhall2020ECCV} for text-guided 3D synthesis. In particular, the objective was to ensure that the rendered images, would match the distribution of the sampled photorealistic images conditioned on a specific text prompt. This process of sampling through optimization is referred to as Score Distillation Sampling (SDS). However, naively applying SDS for 3D synthesis poses several challenges such as over-smoothing, saturated colors as well as Janus-like issues.
Concurrently, ProlificDreamer~\cite{Wang2023NEURIPS} proposed to address these issues with Variational Score Distillation (VSD). 
The key difference between SDS and VSD is that the latter treats a 3D scene as a random variable, as opposed to a single data point. While \cite{Wang2023NEURIPS} addresses some of the issues of SDS, it still suffers from over-saturated colors. In contrast, our work mitigates these challenges by training a 3D-aware generator that directly models the distribution of a diffusion model.

Our work is related to approaches that generate 3D objects conditioned on a single input image. Among the first works in this direction were \cite{Watson2023ICLR, Gu2023ICML, Zhou2023CVPR, Anciukevicius2023CVPR, Karnewar2023ICCV} that proposed training a 3D-aware diffusion model for novel view synthesis. Despite their competitive results, they could only be evaluated on objects from categories seen during training. To mitigate this, an alternative line of research explored using pre-trained 2D diffusion models \cite{Melas2023CVPR, Deng2023CVPR}. To learn control over the camera viewpoint, recently, Zero-1-to-3~\cite{liu2023zero} fine-tuned a pre-trained image-to-text diffusion model~\cite{Rombach2022CVPR} on synthetic data~\cite{Deitke2023ARXIV}. 
In a follow-up work, One-2-3-45~\cite{liu2023one} employed the view-dependent diffusion priors of \cite{liu2023zero} to train a multi-view 3D reconstruction pipeline to enable faster inference. Similar to our work Magic123~\cite{qian2023magic123} uses 2D diffusion together with view-dependent diffusion priors \cite{liu2023zero} for generating 3D textured meshes from a single image.

\section{Method}
\label{sec:method}

Given a single image and a text prompt, our goal is to generate high-quality, photorealistic and diverse 3D content by distilling pre-trained diffusion models. 
First, we show how the concept of continuous distribution modeling avoids the quality degradation brought by mode-seeking (\cref{sec:optimization_framework}). Then we introduce CAD, our  framework that distills knowledge from pre-trained diffusion models to a 3D GAN with multi-view consistent rendering (\cref{sec:gan_training}). However, due to the inherent inductive biases embedded in 2D diffusion models, they tend to only generate images from frontal viewpoints and can not describe the full 3D geometry. We further mitigate this issue by introducing a series of novel strategies to sample multi-view and diversified data from the conditional distribution of diffusion models (\cref{sec:view_sampling}). 

\subsection{Representing 3D Distributions}
\label{sec:optimization_framework}

\paragraph{Preliminaries}%
We first review prior work that leverages diffusion priors for 3D generation.
Given a pre-trained text-to-image diffusion model $p_t(\boldsymbol{x}_t \mid y)$ with the noise prediction network $\boldsymbol{\epsilon}_{\text{pretrain}}(\boldsymbol{x}_t, t, y)$,
SDS~\cite{poole2022dreamfusion} tries to optimize a single NeRF~\cite{Mildenhall2020ECCV} with parameters $\theta$, s.t. its rendered results $\boldsymbol{x}$, given a camera pose $\bc$ sampled from a camera distribution $p_{\bc}(\cdot)$, could minimize the following objective:
\begin{equation}
    \label{equ:sds}
    L_{\mathrm{SDS}} =\mathbb{E}_{t, c}[ D_{\mathrm{KL}}(q_t^{\theta}(\boldsymbol{x}_t \mid \bc) \| p_t(\boldsymbol{x}_t \mid y))],
\end{equation}
where $y$ is the text prompt and $\boldsymbol{x}_t$ is constructed by adding Gaussian noise $\bepsilon$ to $\boldsymbol{x}$ according to a specific timestep $t$ and variance schedules. Note that the gradient of \cref{equ:sds} could be approximated by calculating the noise discrepancy as:
\begin{equation}
\nabla_\theta L_{\mathrm{SDS}}
= \nabla_\theta \mathbb{E}_{t, \boldsymbol{\epsilon}, c}\left[\omega(t)\|\boldsymbol{\epsilon}_{\text {pretrain}}\left(\boldsymbol{x}_t, t, y\right)-\boldsymbol{\epsilon}\|^2\right],
\label{equ:lsds_grad}
\end{equation}
where $\omega(t)$ is a weighting function. Notably, \cref{equ:lsds_grad} resembles the diffusion training loss, thus intuitively SDS could be explained as optimizing the NeRF renderings to look similar to samples generated from a pre-trained diffusion model.

The inherent mode-seeking characteristic of SDS~\cite{poole2022dreamfusion} often leads to sub-optimal generation quality. To address this, VSD~\cite{Wang2023NEURIPS} proposed to replace the single NeRF parametrization $\theta$ from \cref{equ:sds} with a 3D distribution $\mu(\theta \mid y)$ as follows:
\begin{equation}
\label{equ:vsd}
    L_{\mathrm{VSD}} =\mathbb{E}_{t, \bc}[ D_{\mathrm{KL}}(q_t^{\mu}(\boldsymbol{x}_t \mid \bc) \| p_t(\boldsymbol{x}_t \mid y))],
\end{equation}
%
where $\mu(\theta \mid y)$ is parametrized with a set of NeRF instances, referred to as \textit{particles}. Additionally, VSD relies on a dynamically fine-tuned Low-Rank Adaptation (LoRA)~\cite{Hu2022ICLR} to capture the conditional rendering distribution across different camera poses, thus enabling a similar optimization framework under the guidance of denoising score. 


\paragraph{3D Distribution Modeling}%
Turning from single NeRF optimization to distribution modeling is a key for improved 3D synthesis.
However, due to the large computation and memory requirements of VSD~\cite{Wang2023NEURIPS}, the number of NeRF instances that VSD could jointly optimize is quite small ($<10$), hence the 3D distribution modeled by \cite{Wang2023NEURIPS} remains discrete and unexpressive. In practice, we observe VSD still suffers from similar issues as SDS to a certain degree.

Instead, we propose to leverage a generator $G(\boldsymbol{z})$ to capture the continuous 3D distribution. Considering the training efficiency and stability, we follow common practice \cite{Chan2021CVPR, Gao2022NEURIPS, Bahmani2023ICCV} and implement our generator using a StyleGAN2~\cite{Karras2020CVPR} backbone paired with triplane features. In particular, starting from a latent code $\bz \in \mathbb{R}^{d_z}$ drawn from a unit Gaussian distribution, our generator network generates a triplane feature representation $\bP$ as follows:
\begin{equation}
    \begin{aligned}
      G: \mathbb{R}^{d_z}  \rightarrow \mathbb{R}^{ 3 \times N \times N \times C}, \quad
      G(\bz) \rightarrow \bP,
   \end{aligned}
\end{equation}
where $N$, $C$ correspond to the spatial resolution and  channel size respectively.
Note that using a generator to approximate the conditional distribution of the diffusion model fundamentally resolves mode-seeking issues. Additionally, instead of learning the triplane features directly from 
$\bz$, the generator also incorporates a non-linear mapping network, which maps $\bz$ to latent vectors $\bw \in \mathcal{W}$, controlling the generator
through adaptive instance normalization (AdaIN)~\cite{Huang2017ICCV} at each convolution layer. The intermediate latent space $\mathcal{W}$ shares many desirable properties, enabling us to perform continuous 3D interpolation and single-view reconstruction by inversion.




\subsection{Consistent Adversarial Distillation}
\label{sec:gan_training}

\paragraph{Adversarial Distillation}%
It should be noted the actual optimization goal of score distillation is
\begin{equation}
\label{equ:x_0}
    \min _\mu D_{\mathrm{KL}}\left(q_0^\mu\left(\boldsymbol{x}_0 \mid c\right) \| p_0\left(\boldsymbol{x}_0 \mid y\right)\right).
\end{equation}
SDS tackles \cref{equ:x_0} by breaking it down into multiple sub-optimization problems, each associated with a unique diffused distribution indexed by $t$, as indicated in \cref{equ:sds} and \cref{equ:vsd}. However, doing optimization using noisy discrepancy can easily lead to noticeable quality degradation in comparison to the direct denoising sampling approach, as reported in \cite{poole2022dreamfusion, Wang2023NEURIPS}.


%
To achieve high-quality and diversified generation, we introduce adversarial distillation shown in \cref{fig:architecture}, an innovative approach for optimizing the 3D generator $G(\cdot)$ s.t. its sampled data match the samples from a pre-trained diffusion model, $p_0\left(\boldsymbol{x}_0 \mid y\right)$, as follows:
\begin{equation}
    \label{equ:G_p0}
    \min _G D_{\mathrm{KL}}(\colorbox{lightblue}{$p^G(\boldsymbol{x} \mid \bz, \bc)$} \| p_0\left(\boldsymbol{x}_0 \mid y\right)),
\end{equation}
where $\boldsymbol{x}$ is the rendered image from triplane $\mathbf{P}=G(\bz)$ under camera pose $\bc$.
Instead of directly computing the KL-divergence, we solve \cref{equ:G_p0} by learning an implicit likelihood model, namely a GAN~\cite{Goodfellow2014NIPS}.

In particular, we train our model using an adversarial objective and adopt the non-saturating GAN loss with R1 regularization \cite{Mescheder2018ICML} from \cite{Karras2019CVPR} as follows:
\begin{equation}
\begin{aligned}
\label{equ:nsgan}
    L_{\text {D}} = &\mathbb{E}_{\bz \sim p_z, \boldsymbol{c} \sim p_{\boldsymbol{c}}}
    \big[f\left(D(\mathcal{R}(G(\bz), \bc)\right))\big] \,+ \\
    &\mathbb{E}_{I \sim p_0}\Big[f\left(-D(I)\right)+\lambda \| \nabla D(I)\|^2\Big]
\end{aligned}
\end{equation}
\vspace{-2em}
\begin{align}
    L_{\text {G}} = -\mathbb{E}_{\bz \sim p_z, \boldsymbol{c} \sim p_{\boldsymbol{c}}}\left[f\left(D\left(\mathcal{R}(G(\bz), \bc)\right)\right)\right],
\end{align}
where $f(u)$ is defined as $f(u)= -\log(1+ \exp(-u))$, $I$ is sampled from the diffusion model, $\mathcal{R}(\cdot)$ denotes the volumetric renderer used to render the generator's output that parametrizes the fake data distribution, $D(\cdot)$ denotes the discriminator and $\lambda$ is a hyperparameter.

To render an image, for every coordinate along the camera ray, we sample features from the triplanes $\bP$ and aggregate them with summation. Next, a light-weight MLP decoder $M(\cdot)$ will explain the aggregated features $\mathbf{f} \in \mathbb{R}^{d_{\mathrm{f}}} $ into radiance $\bc \in \mathbb{R}^3$ and volume density $\sigma \in \mathbb{R}^{+}$:
\begin{equation}
\begin{aligned}
      M: \mathbb{R}^{d_{\mathrm{f}}}  \rightarrow \mathbb{R}^{3} \times \mathbb{R}^{+}, \quad
      M(\mathbf{f}) \rightarrow (\bc, \sigma).
   \end{aligned}
\end{equation}
We follow common practice and use classical volumetric rendering equation with hierarchical sampling~\cite{Mildenhall2020ECCV} to predict the final RGB colors for every pixel in the image. Unlike score distillation, our distillation scheme is directly based on high-quality and clean samples drawn from the diffusion model, which effectively overcomes over-saturation issues. Furthermore, since adversarial distillation does not minimize the pixel-wise distance to intermediate images or denoising scores that are usually incoherent, our method can reliably generate photorealistic texture details.

\begin{figure}
    \centering
    \includegraphics[width=\linewidth]{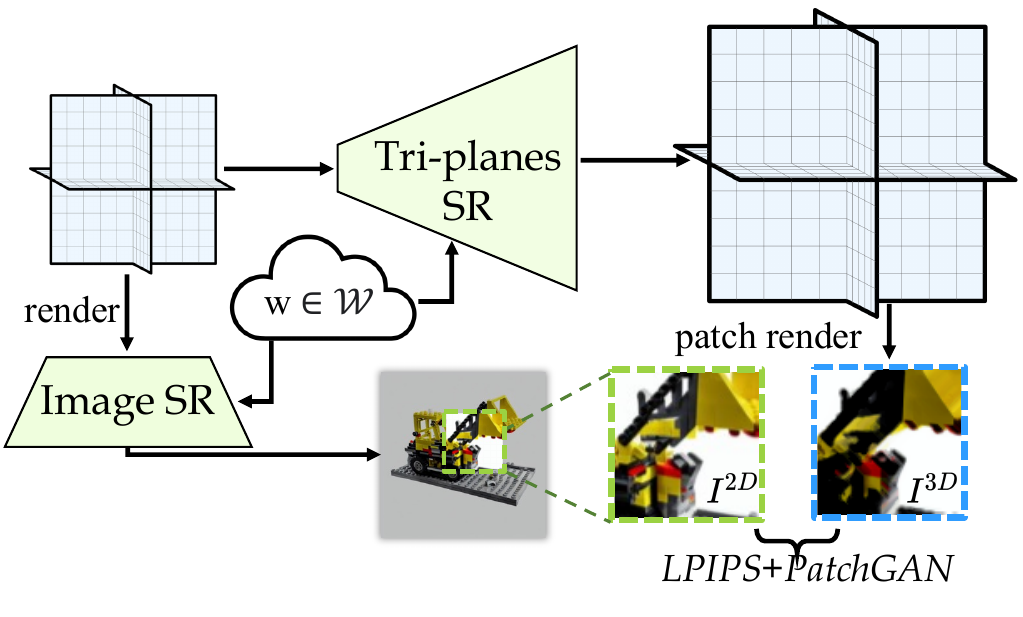}
    \vspace{-2.5em}
    \caption{We train a 3D consistent GAN by baking the 2D upsampler to a 3D upsampler through minimizing the patch-level differences between the renderings of the two branches.}
    \label{fig:dual_branch}
    \vspace{-1.2em}
\end{figure}

\paragraph{Consistency}%
Directly training a 3D GAN at high resolutions using volume rendering is computationally infeasible. Therefore, existing methods \cite{chan2022efficient, Shen2023CVPR, Bahmani2023ICCV} rely on image-space convolutions to upscale the resolution of the raw renderings. While 2D upsampling can ensure high-quality and 3D consistent renderings with imperceptible artifacts for forward-facing data \cite{chan2022efficient}, we experimentally observe that naively applying this approach to 360° object-level 3D synthesis leads to significant multi-view inconsistencies.

As shown in \cref{fig:dual_branch}, we resolve this issue by baking the 2D upsampler branch into a 3D triplane upsampler via patch-wise similarity following~\cite{Chen2023ICCVb}. The motivation here is two-fold: 1) By sharing the same latent code $\bw$, the image rendered through the 3D upsampler has the potential to capture high-frequency details similar to those produced by the 2D branch output. 2) Once trained, renderings from the same triplane trivially preserve 3D consistency. Therefore, given a latent code $\bz$ and a camera pose $\bc$, we use the LPIPS~\cite{zhang2018perceptual} loss to optimize the 3D upsampler as:
\begin{equation}
\label{equ:mimic}
    L_{\text{consistency}}=\operatorname{LPIPS}\left(I^{3 D}, \operatorname{sg}\left(I^{2 D}\right)\right),
\end{equation}
where $I^{2 D}$ and $I^{3 D}$ are images generated from the 2D and the 3D upsampling branches respectively and $\operatorname{sg}$ denotes stopping gradient. For efficiency we only calculate \cref{equ:mimic} in $64^2$ patch-level. We also train a small patch discriminator to ensure the renderings of upsampled triplane preserving high-frequency details, using a similar objective as in \cref{equ:nsgan}.


\begin{table}
    \centering
    \resizebox{1.0\linewidth}{!}{%
    \begin{tabular}{l|cccc}
    \hline
     & \multicolumn{2}{c}{Image Score $(\uparrow)$} & \multicolumn{2}{c}{Text Score  $(\uparrow)$}\\
    Method & ViT-L/14   & ViT-B/32 & ViT-L/14 & ViT-B/32 \\ \hline
    DreamFusion & 65.71 & 72.61 & \cellcolor{tabsecond}24.53 & 28.81 \\
    Zero-1-to-3  & 68.10 & \cellcolor{tabthird}77.42 & 21.35 & 27.90 \\
    ProlificDreamer &  \cellcolor{tabthird}72.72 & 74.46 &  \cellcolor{tabthird}24.51 &\cellcolor{tabthird}29.39 \\
    Magic123  & \cellcolor{tabsecond}75.61 & \cellcolor{tabsecond}84.02 &  23.95 &  \cellcolor{tabfirst}\bf{29.89} \\
    Ours  &  \cellcolor{tabfirst}\bf{82.56} &  \cellcolor{tabfirst}\bf{89.16} &  \cellcolor{tabfirst}\bf{25.32} & \cellcolor{tabsecond}29.81 \\
    \hline
    \end{tabular}
    }
    \vspace{-0.5em}
    \caption{\textbf{Quantitative Evaluation.} We measure the CLIP similarity score~\cite{Radford2021ICML}. For the Image Score we compute the CLIP distance between rendered and reference views, while for the Text Score, we compute the CLIP distance between rendered and text promots. We color each row as \colorbox{tabfirst}{best}, \colorbox{tabsecond}{second best}, and \colorbox{tabthird}{third best}.}
    \label{tab:clip_results}
    \vspace{-1.2em}
\end{table}

\subsection{Multi-View Sampling}
\label{sec:view_sampling}

In this section, we discuss our strategy for sampling and filtering high-quality images from $p_0(\cdot)$ in order to effectively perform adversarial distillation and calculate \cref{equ:nsgan}.

\paragraph{Sampling}%
Diffusion models \cite{Saharia2022NEURIPS, Rombach2022CVPR} can produce high-quality images, but often exhibit a bias towards generating frontal-facing images, thus failing to adequately span the entire azimuth $\phi \in [0, 2\pi]$ and elevation $\theta \in [0, \pi]$ angles. However, this bias can make the discriminator overfit to particular viewpoints, resulting in inaccurate gradients when optimizing the 3D generator. To resolve this we leverage the view-dependent diffusion model of Zero-1-to-3~\cite{liu2023zero} to produce more diverse views given a single input image.


\paragraph{Pruning}%
Although the view-conditioned diffusion model could assist us towards obtaining free-view diffusion priors in 360°, we still observe several issues. For instance, the generated viewpoints may not adhere to the target camera pose, or there may be substantial geometric degradation in the generated images, resulting in visible 3D misalignment compared to the reference image as shown in \cref{figure:prune_visualize}. Although the discriminator can tolerate some errors to a certain extent, the frequent occurrence of misalignment will significantly undermine the stability of GAN training.


To this end, we propose a camera pose pruning strategy aiming at filtering out ``bad diffusion samples''. Specifically, given a reference image and a target camera pose, we synthesize $N$ samples in parallel with reverse denoising. To capture the essential geometric characteristics of the target pose, we compute a shared overlapping mask $m$ according to the generated samples, which is then dilated to allow a certain degree of geometry deformation. We consider the pixels within the mask as ``good'', and compute the maximum number of ``bad pixels'' (\ie pixels outside the mask) among the generated samples. If this maximum count exceeds a predefined threshold, we discard this viewpoint. 

In addition to the geometric consistency check, we also assess the semantic and structural similarity between sampled $I_{\text{syn}}^{i}$ and reference views $I_r$ using the pre-trained CLIP image encoder as follows:
\begin{align}
    \min_{i=1:N}\texttt{CLIP}_{\text{image}}(I_{\text{syn}}^{i}, I_r).
    \label{equ:clip_score}
\end{align}
If the score of \eqref{equ:clip_score}, for a specific view, is below a threshold, we reject this viewpoint. Viewpoints that fail to pass any of these two checks are discarded. Among images from the same view, we select the sample with the highest CLIP score.

\begin{figure*}[t!]
    \begin{center}
        \includegraphics[width=1.0\linewidth]{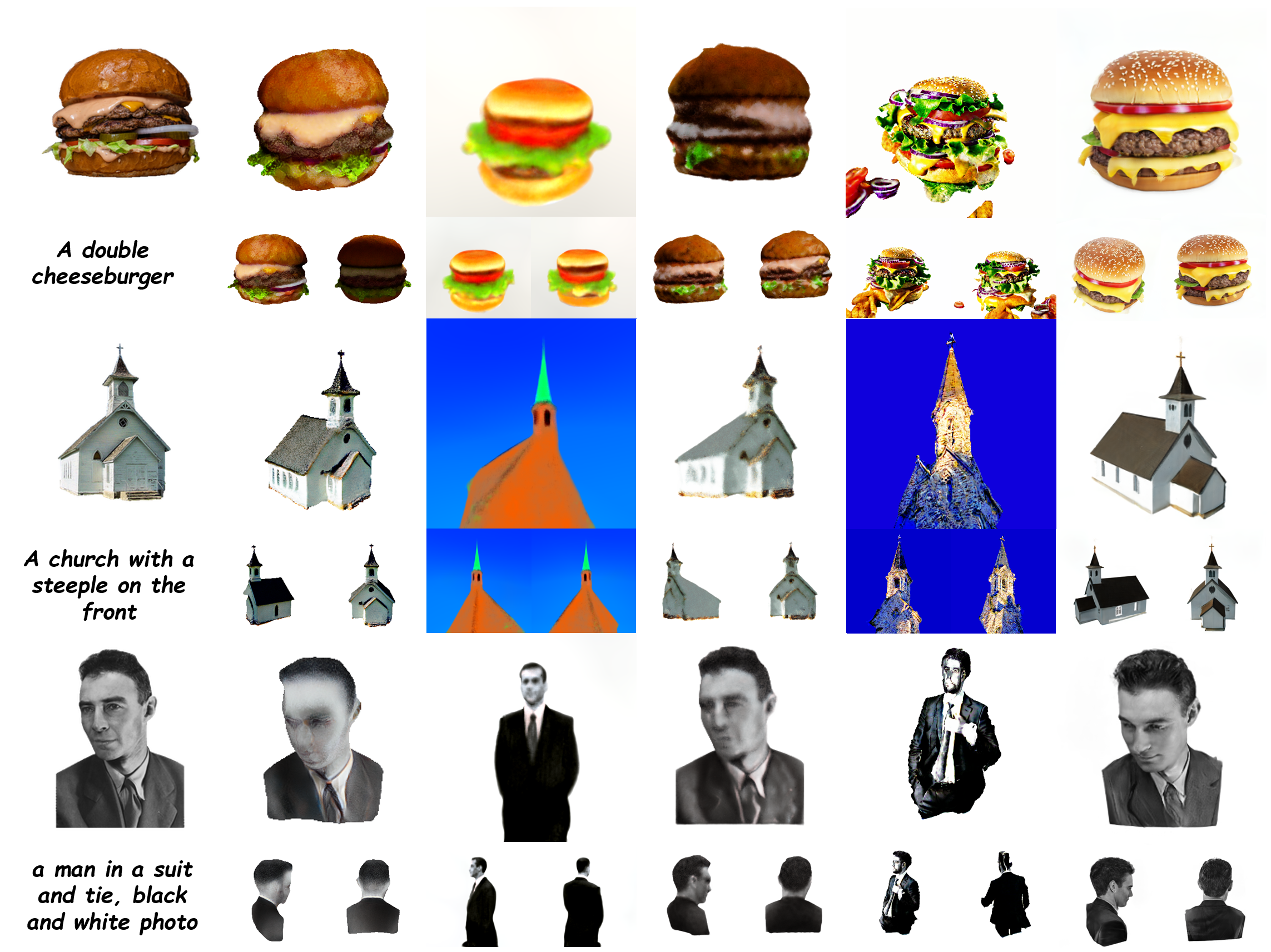}
        \begin{tabularx}{1.0\linewidth}{YYYYYY}
             \vspace{-1.3em} Reference & \vspace{-1.3em} Magic123~\cite{qian2023magic123} & \vspace{-1.3em} DreamFusion~\cite{poole2022dreamfusion} & \vspace{-1.3em} Zero-1-to-3~\cite{liu2023zero} & \vspace{-1.3em} ProlificDreamer~\cite{Wang2023NEURIPS} & \vspace{-1.3em} Ours
        \end{tabularx}
    \end{center}
    \vspace{-2.0em}
    \caption{\textbf{Qualitative Evaluation.} Our method yields more photo-realistic 3D generations conditioned on a single image (left-most column), with significantly less artifacts compared existing SDS-based pipelines \cite{poole2022dreamfusion, Wang2023NEURIPS}. In comparison to the single-view conditioned 3D generation approaches \cite{qian2023magic123, liu2023zero}, our generations have significantly fewer artifacts and look more realistic.}
    \label{figure:qualitative_compare}
    \vspace{-1.2em}
\end{figure*}

\paragraph{Distribution Refinement}%
Although the generated images, produced by the view-dependent Zero-1-to-3~\cite{liu2023zero} are more diverse in terms of camera poses, they often lack appearance variations and exhibit blurred textures. To mitigate this, we propose to further refine the prior extracted by \cite{liu2023zero} with a powerful text-guided 2D diffusion model, such as DeepFloyd~\cite{deepfloyd}. Specifically, for the sample generated with \cite{liu2023zero}, we add noise over the image with specific strength and then denoise it by~\cite{deepfloyd} to generate high-quality and diversified refinement, which still retains similar pose and semantics with the input. In general, using a higher noise level typically yields
more diverse generations, however this also poses a challenge in maintaining the original pose information. Therefore, we also try ControlNet~\cite{Zhang2023ICCV} conditioned on depth maps to achieve better 3D diversity while preserving pose information. Please check more analysis in \cref{Analysis}.

\section{Experimental Evaluation}
\label{sec:results}

\paragraph{Datasets}%
In our evaluation, we consider images from three sources: (i) high-quality real-world images from the Internet, (ii) synthetic scenes from Blender including \emph{Synthetic-NeRF}~\cite{Mildenhall2020ECCV} and \emph{Synthetic-NSVF}~\cite{Liu2020NeuRIPS}, and (iii) images generated by a text-to-image diffusion model. For the case of objects from \cite{Mildenhall2020ECCV, Liu2020NeuRIPS}, we only utilize a single image and ignore its associated pose information. When no text descriptions are provided, we employ an off-the-shelf image caption pipeline~\cite{Li2022ICML} to obtain a per-image text description.


\paragraph{Baselines} We assess the performance of our approach against two significant image-to-3D methods including Magic123~\cite{qian2023magic123} and Zero-1-to-3~\cite{liu2023zero}, as well as two notable text-to-3D techniques, DreamFusion~\cite{poole2022dreamfusion} and ProlificDreamer~\cite{Wang2023NEURIPS}. We use the improved implementation from ThreeStudio~\cite{threestudio2023} for the baseline training with shared input view, prompt, pose and resolution across all the methods.

\begin{figure*}[t!]
    \begin{center}
        \includegraphics[width=1.0\linewidth]{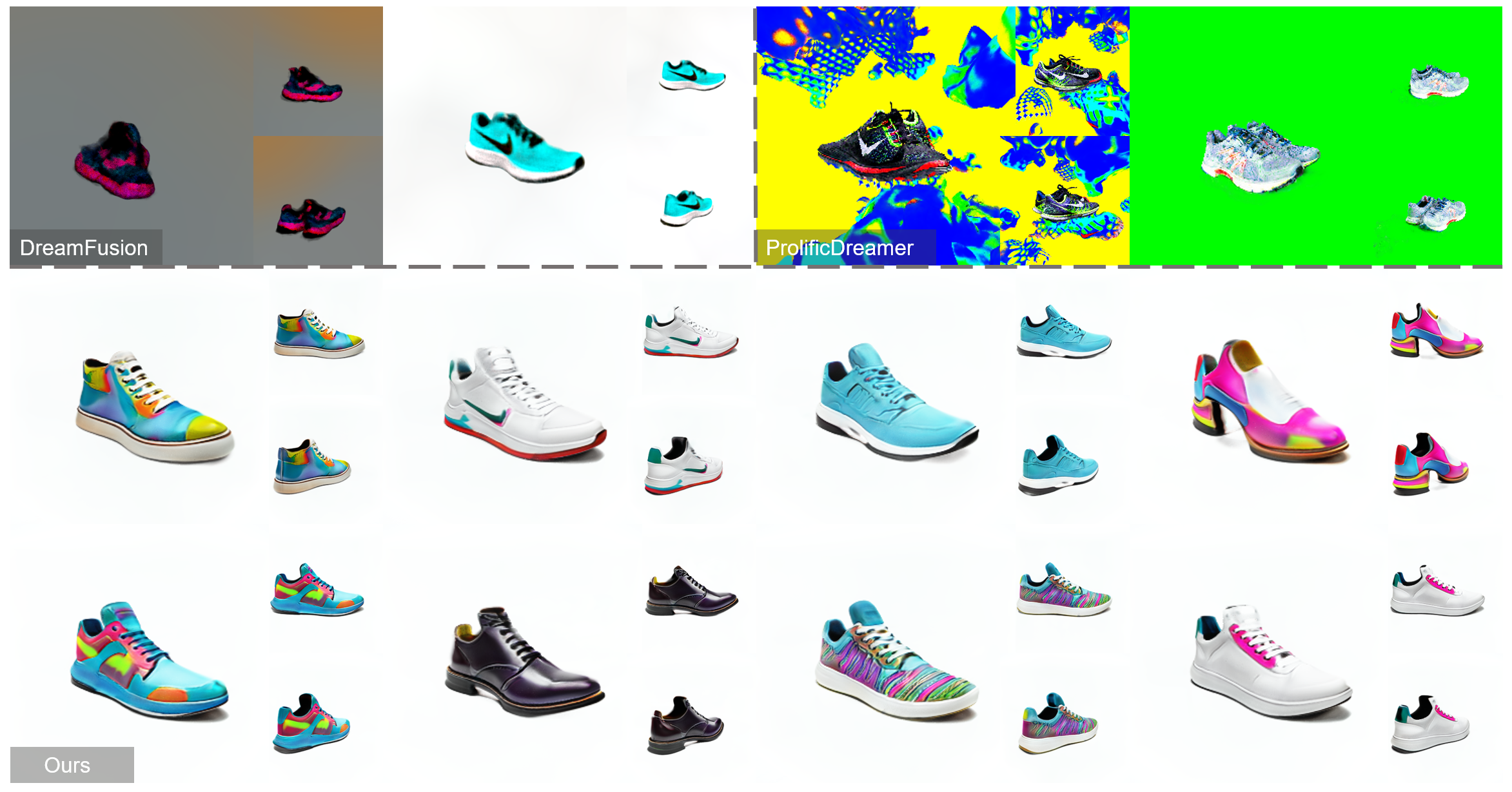}
    \end{center}
    \vspace{-1.8em}
    \caption{\textbf{Diversity and quality comparisons with the baselines.} By distilling the diffusion prior into a 3D generator, our method is better in terms of diversity, quality and photorealism. Note that in contrast to DreamFusion~\cite{poole2022dreamfusion} and ProlificDreamer~\cite{Wang2023NEURIPS} that require re-optimizing for more than $10$ hours for each novel sample, our model can generate diverse and novel 3D objects within a second. }
    \label{figure:diversity_compare}
    \vspace{-1.2em}
\end{figure*}

\paragraph{Evaluation Metrics}%
We follow common practice \cite{poole2022dreamfusion, liu2023one, qian2023magic123} and report the CLIP similarity score~\cite{Radford2021ICML} with different image-based ViT~\cite{Dosovitskiy2021ICLR} and text-based architectures trained by OpenAI. Specifically, we create a 360° camera trajectory orbiting the object with 120 frames that cover different views. For each view, we calculate the image and text similarity with a reference image and a text prompt respectively. In our evaluation, we consider a total of $1,200$ images gathered from our data sources. Additional details as well as a user study that evaluates the subjective synthesis quality and diversity are provided in the supplementary.

\paragraph{Implementation Details}%
In our framework, we adopt the 3D generator architecture of EG3D~\cite{chan2022efficient} with a few key modifications tailored to our specific 3D adversarial distillation formulation. First, we remove the pose conditioning from the generator, as we are interested in modeling generic objects.
To guide the generator to learn the correct 3D pose prior, we follow \cite{chan2022efficient} and inject the absolute camera pose into the discriminator.
During optimization we generate 10K samples per object and enable the adaptive discriminator augmentation (ADA)~\cite{Karras2020NEURIPS} to stabilize the adversarial training. During the distribution refinement step, we employ a noise strength of $0.8$ for the depth-conditioned Control-Net~\cite{Zhang2023ICCV} and
noise strengths of $0.3$ and $0.7$ for the low and high-resolution branches of DeepFloyd~\cite{deepfloyd}. Moreover, we leverage the view-dependent prompting to avoid the multi-face issue while dealing with asymmetric objects.
To obtain a 3D distribution based on the image and prompt, the training process takes approximately 3
days for images of $256^2$ resolution, where training the 2D upsampler branch requires 1.5 days and fine-tuning together with the 3D upsampler requires another 1.5 days on 4$\times$V100 GPUs. Additional implementation details are provided in the supplementary.


\subsection{Comparisons}
We now compare our adversarial distillation framework  with existing score distillation methods and demonstrate that it performs better in terms of generation quality and diversity.

\paragraph{Quantitative Comparison}%
In \tabref{tab:clip_results}, we compare all methods \wrt their consistency in terms of appearance and semantic similarity. In particular, we note that our method significantly outperforms all baselines in terms of the CLIP image score. This is expected, as our renderings are more photorealistic and do not suffer from over-saturation and over-smoothing issues apparent in existing SDS-based pipelines. In terms of text similarity metrics, our approach again outperforms all baselines when using the ViT-L/14 pre-trained model, while performing on-par with \cite{qian2023magic123} when using the ViT-B/32 pre-trained model. The superiority of our model in both metrics indicates its ability to better match the conditioning signal and generate more high-quality 3D contents.

\paragraph{Qualitative Comparison}%
In \figref{figure:qualitative_compare}, we qualitatively compare our model with several baselines. 
Compared to Dreamfusion~\cite{poole2022dreamfusion} that generates blurry 3D objects with highly saturated colors, our model produces more photorealistic objects with fine details. ProlificDreamer~\cite{Wang2023NEURIPS} further proposes VSD, which replaces a single NeRF fitting into distribution matching, demonstrating improved texture details. However, VSD tends to generate multi-face 3D like shown in the second row of \figref{figure:qualitative_compare}. Even for the symmetric objects like burger, its 3D rendering exhibits non realistic colors similar to \cite{poole2022dreamfusion}. In contrast, both Magic123~\cite{qian2023magic123} and Zero-1-to-3~\cite{liu2023zero} partially mitigate the color shift issues, but still generate objects with blurry appearance. Instead, our adversarial distillation framework attains a higher level of photorealism, while preserving the multi-view consistent fine details.

\paragraph{Diversity}%
A key benefit of our approach is that it can generate diverse 3D objects after distillation. To showcase this, in this section, we compare our model with DreamFusion~\cite{poole2022dreamfusion} and ProlificDreamer~\cite{Wang2023NEURIPS} in terms of their ability to produce diverse renderings conditioned on the text prompt ``\textit{running shoes}''. For our model, we condition our generation on the same text prompt and an input image showing a white shoe, which we provide in the supplement. Note that currently ProlificDreamer doesn't share the multi-particle implementation thus to obtain the diversity we have to use different seeds for re-optimization. As shown in \figref{figure:diversity_compare}, even with variations, both DreamFusion~\cite{poole2022dreamfusion} and ProlificDreamer~\cite{Wang2023NEURIPS} produce smooth objects with highly-saturated colors that exhibit Janus-like issues. By contrast, our method obtains significantly better quality and diversity. Moreover, since the the learned 3D generator models the target distribution, our model could generate diverse and novel 3D objects significantly faster than all the baselines without re-optimization.

\begin{figure}[!t]
    \begin{center}
    \includegraphics[width=1.0\linewidth]{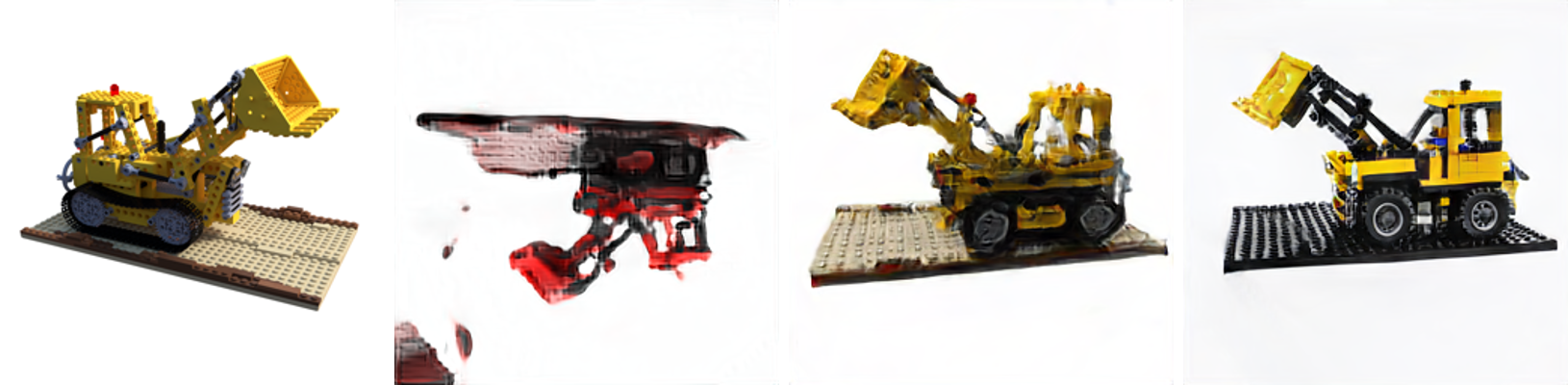}
        \begin{tabularx}{1.0\linewidth}{YYYY}
            Reference & $-$ Pruning & $- \text{Pruning}^{\dagger}$ & $+$ Pruning
        \end{tabularx}
    \end{center}
    \vspace{-1.5em}
    \caption{\textbf{Impact of Pose Pruning.} The first column, shows the reference image, the second shows the output when no pose pruning is used, the third shows a variant of our model without pose pruning and ADA~\cite{Karras2020NEURIPS} but with a larger number of diffusion samples 100K and the last shows the output using the proposed pose pruning.}
    \label{figure:ablation_prune}
    \vspace{-0.5em}
\end{figure}

\subsection{Ablation Study}

\paragraph{Effectiveness of Pose Pruning}%
To demonstrate the importance of the camera pose pruning (see \secref{sec:view_sampling}), we investigate its
impact on the Lego scene from Synthetic-NeRF~\cite{Mildenhall2020ECCV}. We start by removing the camera pose pruning and observe that the the rendering quality deteriorates significantly, as shown in the second column of \figref{figure:ablation_prune}. To prevent this behavior from being caused by the information leakage of Adaptive Discriminator Augmentation (ADA)~\cite{Karras2020NEURIPS}
, we further increase the diffusion samples from 10K to 100K while also removing ADA \cite{Karras2020NEURIPS}. In this scenario, the pose and color seem plausible but the generation quality is still relatively low. In contrast, using camera pose pruning enables the training of a superior 3D GAN, even with limited amount of data, highlighting its efficacy and critical role in enhancing the overall performance of adversarial distillation.
 
\begin{figure}[!t]
    \begin{center}
    \includegraphics[width=1.0\linewidth]{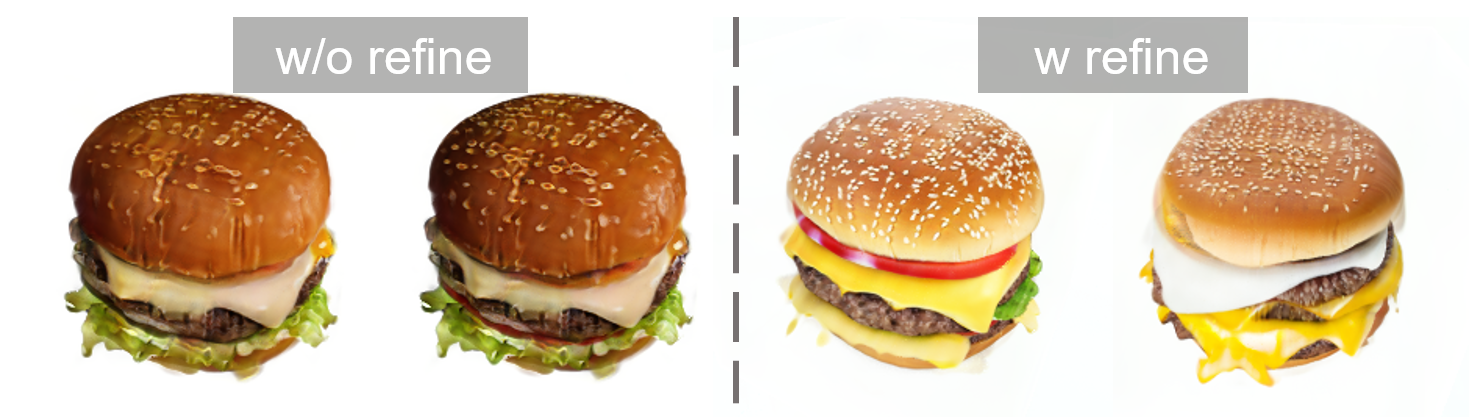}
        \begin{tabularx}{1.0\linewidth}{YYYY}
            Seed 1 & Seed 2 & Seed 1 & Seed 2
        \end{tabularx}
    \end{center}
    \vspace{-1.5em}
    \caption{\textbf{Impact of Distribution Refinement.} Removing the Distribution Refinement results in inferior diversity and quality.  }
    \label{figure:ablation_refine}
    \vspace{-0.5em}
\end{figure}

\paragraph{Distribution Refinement}%
Distribution refinement is a key step to ensure the generation quality and diversity of our distillation pipeline. As shown in \figref{figure:ablation_refine}, optimizing the 3D generator using the original samples from the view-dependent Zero-1-to-3~\cite{liu2023zero} without enough diversities could learn to capture a basic shape and appearance but fails to produce high-quality and diverse generations.

\begin{figure}[!t]
    \begin{center}
    \includegraphics[width=1.0\linewidth]{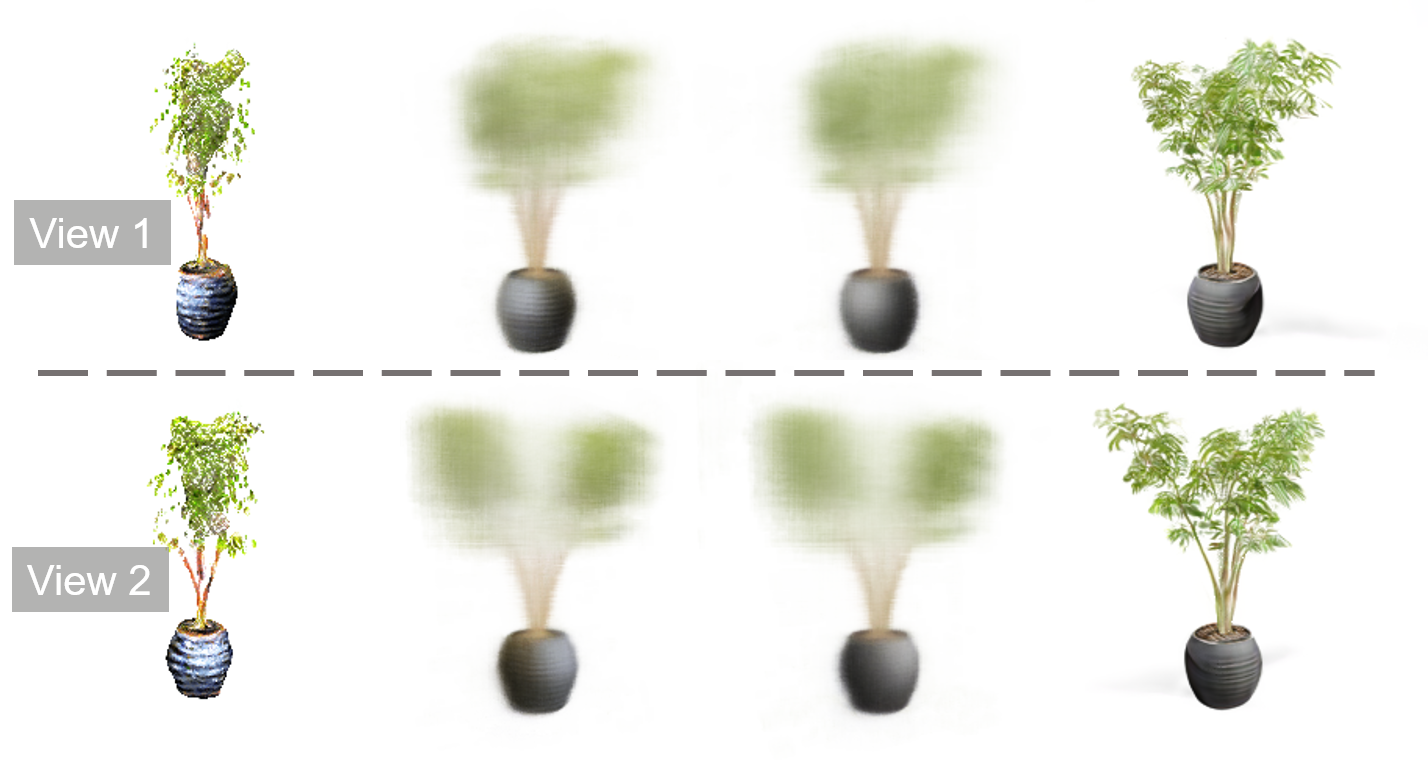}
        \begin{tabularx}{1.0\linewidth}{YYYY}
            Magic123~\cite{qian2023magic123} & One Mode & $\text{One Mode}^{\dagger}$ & Ours
        \end{tabularx}
    \end{center}
    \vspace{-1.5em}
    \caption{\textbf{Impact of Single Mode Fitting and Distribution Modeling.} We compare our model with Magic123~\cite{qian2023magic123}, as well as with two NeRF variants optimized using the raw sampled images ~\cite{liu2023zero} (second column) and the refined samples (third column). Our model could yield realistic and 3D consistent renderings.}
    \label{figure:gan_modeling}
    \vspace{-0.5em}
\end{figure}

\paragraph{Single Mode v.s. Distribution Modeling}%
 We would like to further demonstrate the importance of distribution matching to achieve photorealistic and diversified 3D generation. We now consider the Ficus scene from Synthetic-NeRF~\cite{Mildenhall2020ECCV}. This scene is particularly challenging as it contains thin structures with high-frequency texture details. As shown in \figref{figure:gan_modeling}, Magic123~\cite{qian2023magic123}, which tries to find a single mode under the guidance of score direction, shows poor geometry and rendering quality. Moreover, we also try to fit a single NeRF~\cite{Mildenhall2020ECCV} directly using the raw sampled images from Zero-1-to-3~\cite{liu2023zero} without any refinement (see second column) and the improved samples using proposed distribution refinement strategy (see third column). As we could see, both these variants yield low-quality rendering with blurry artifacts. In contrast, our method CAD could render realistic and highly detailed frames with multi-view consistency.

\section{Conclusion}

In this paper, we proposed CAD, a new approach for generating high-quality, photoreaslisitc and diverse 3D objects conditioned on a single image and a text prompt. We showcase that by directly modeling the distribution gap between multi-view renderings and diffusion priors, our pipeline addresses several issues related to SDS-based pipelines. Despite the promising results of our model, it still has several limitations. One of the main bottlenecks of our framework is the optimization speed, which is hindered by the inherently slow process of volumetric rendering required to generate high-resolution frames ($\sim$ 0.1 FPS). A potential solution to address this limitation is the adoption of more efficient rendering techniques, such as Gaussian Splatting~\cite{Kerbl2023SIGGRAPH}. Moreover, currently we only consider a single conditioning input, joint training with multiple conditions potentially could result in more diverse geometry and appearance variations. Finally, although we mainly focused on the object-level 3D synthesis, extending CAD to scene-level scenarios is also a very promising research direction.

\noindent\textbf{Acknowledgements: }We appreciate helpful discussions with Guandao Yang, Boxiao Pan, Zifan Shi, Chao Xu, Xuan Wang, Ivan Skorokhodov, Jingbo Zhang, Xingguang Yan and Connor Zhizhen Lin. The work described in this paper was substantially supported by a GRF grant from the Research Grants Council (RGC) of the Hong Kong Special Administrative Region, China [Project No. CityU11208123]. Despoina Paschalidou is supported by the Swiss National Science Foundation under grant number P500PT 206946.
{
    \small
    \bibliographystyle{ieeenat_fullname}
    \bibliography{main, bibliography_long,bibliography,bibliography_custom}
}

\clearpage

\appendix
\setcounter{page}{1}
\maketitlesupplementary

\section{Overview}

In this supplemental material,  additional implementation details and experimental results are provided, including:
\begin{itemize}
  \setlength\itemsep{0.1em}

  \item Results of the user study (Section \ref{user_study});
  
  \item More details about the whole distillation pipeline (Section \ref{implement});
  
  
  \item More analysis regarding the pose pruning, distribution refinement, diversity and consistency training (Section \ref{Analysis});
  
  \item Video demo, which shows the photorealistic generation of our method and qualitative video comparisons. Please refer to \textcolor{orange}{\textit{video demo}} in the project website.
\end{itemize}

\section{User Study} \label{user_study}
\begin{figure}[t!]
    \begin{center}
    \includegraphics[width=1.0\linewidth]{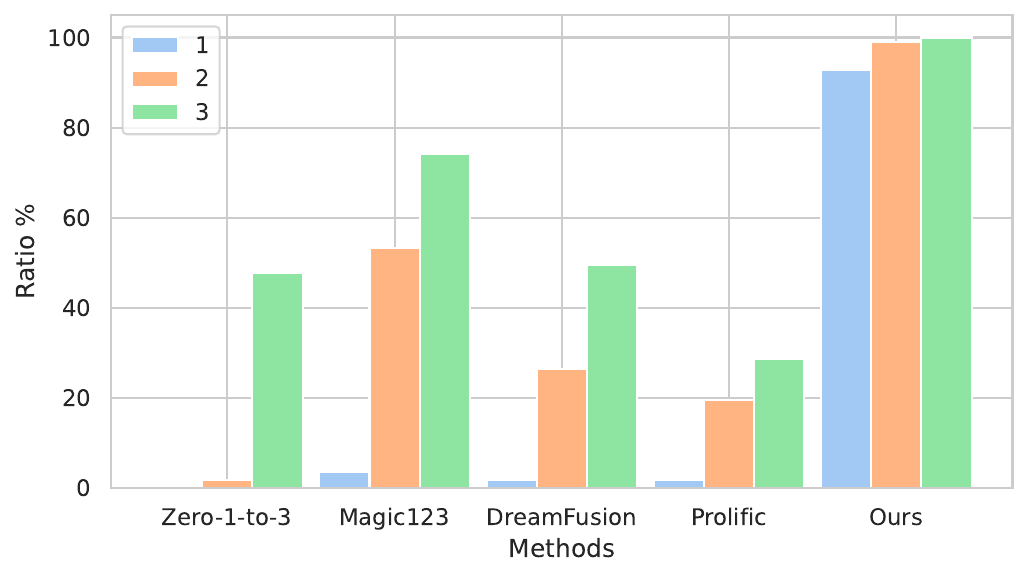}
    \end{center}
    \vspace{-2.0em}
    \caption{\textbf{Distribution of User Preferences}. This graph illustrates the percentages of participatns that ranked each method as their top three choices, indicated by 1, 2, and 3, respectively.}
    \label{figure:user_study}
    \vspace{-1.0em}
\end{figure}

To better evaluate the subjective quality and diversity of 3D generation, we conducted a user study to compare our method with existing baselines. Specifically, we created a test set comprising 10 images and prompts. For each case, we utilized Zero-1-to-3~\cite{liu2023zero} and Magic123~\cite{qian2023magic123} for generating one 3D object, and Dreamfusion~\cite{poole2022dreamfusion}, ProlificDreamer~\cite{Wang2023NEURIPS}, along with our method for synthesizing multiple 3D objects. Participants were then asked to rank the five groups from highest to lowest, based on a comprehensive inspection of various aspects, including rendering quality, photorealism, and generation diversity. We collected surveys from 22 participants and calculated the percentages of each method being selected as the top 1, 2, and 3, with the statistics shown in Figure~\ref{figure:user_study}. Our method demonstrates clear superiority over other methods, with more than 92\% of users selecting it as the best result, significantly surpassing the second-highest percentage of 3.6\% for Magic123~\cite{qian2023magic123}.


\section{Implementation Details} \label{implement}

The 3D generator architecture implemented in our paper is an adaptation of EG3D~\cite{chan2022efficient}, with several critical modifications designed for our 3D adversarial distillation framework: 1) We have omitted the generator pose conditioning since our model targets general 3D objects, not specifically pose-correlated faces. 2) To achieve increased compactness and accelerated optimization, the maximum resolution of the triplanes is reduced to $128^2$. 3) The raw (volumetric) rendering resolution of the triplanes is fixed at $64^2$ throughout the distillation process. We rely on 3D upsampling for maintaining multiview consistency. The generated triplanes are confined within a $[-0.7,0.7]^3$ box, and we utilize a field of view (FOV) of $49.1^{\circ}$, following a pinhole camera model. We incorporate the absolute camera pose into the discriminator to facilitate the generator in learning the correct 3D prior. For the reference viewpoint, we set the azimuth angle ($\phi$) to $180^{\circ}$, the distance from the center $r = 2$, and determine the polar angle ($\theta$) using the pose estimation module from One-2-3-45~\cite{liu2023one}. The pose for any other viewpoint is then derived through relative transformation.

To ensure the efficiency of the adversarial distillation, we do not adopt the on-the-fly way to sample $I\sim p_0$ from a frozen diffusion model, instead we cache the sampling in advance so that during the distillation we could directly fetch the prior from the memory. Although sampling more data prior will can lead to improved quality and diversity, considering the efficiency we only restrict the pre-sampling number to be 10K and enable the adaptive discriminator augmentation (ADA) to stabilize the adversarial training. Each sample is associated with a unique pose, uniformly distributed across a sphere with a radius of $2$.

\begin{figure*}[!t]
    \begin{center}
    \includegraphics[width=1.0\linewidth]{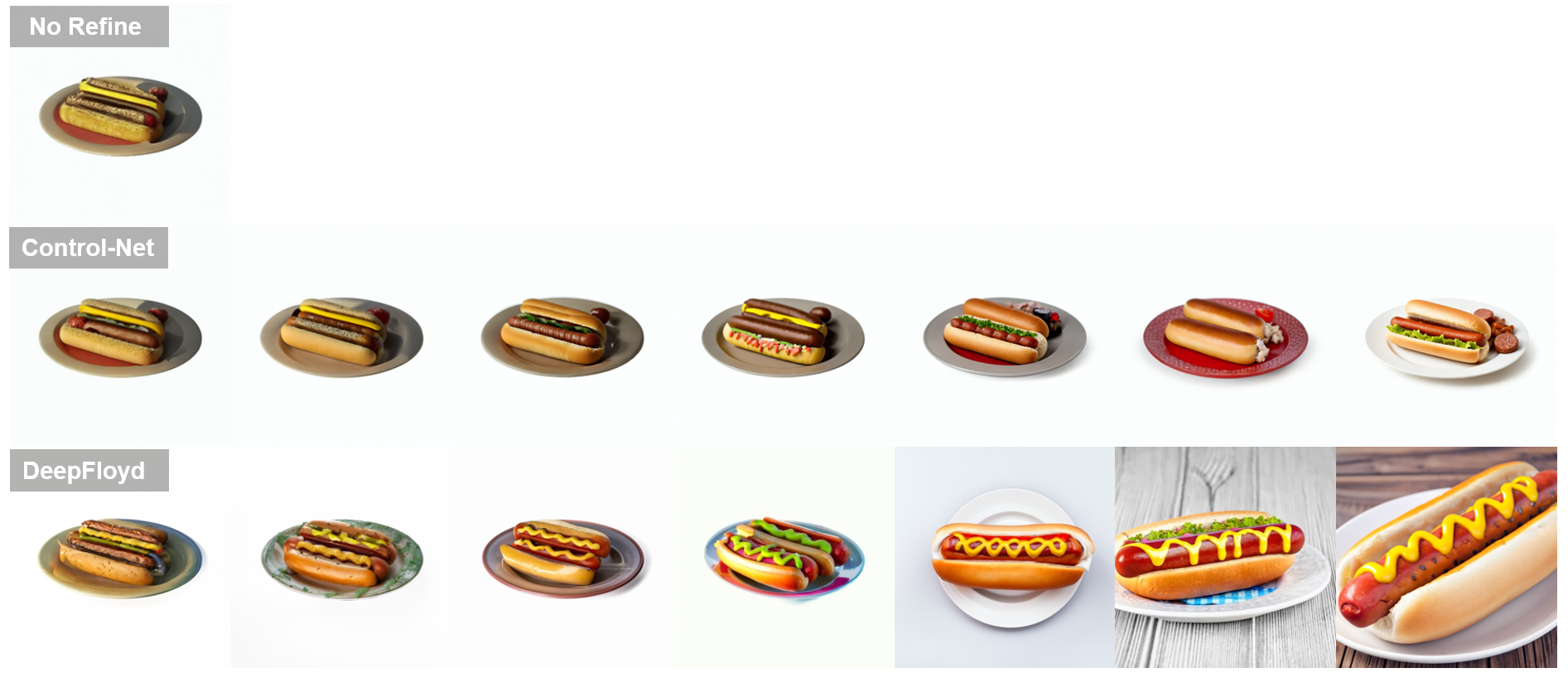}
        \begin{tabularx}{1.0\linewidth}{YYYYYYY}
            $s=0.4$ & $s=0.5$ & $s=0.6$ & $s=0.7$ &$s=0.8$ & $s=0.9$ & $s=1.0$
        \end{tabularx}
    \end{center}
    \vspace{-1.5em}
    \caption{\textbf{Effects of the noise strength.} We show the refinement results by using different diffusion models and various noise strengths.}
    \label{figure:noise_strength}
\end{figure*}

In the process of distribution refinement, we apply different noise strengths to different diffusion priors. Specifically, a noise strength of 0.8 is utilized for the depth-conditioned Control-Net. In the case of DeepFloyd, noise strengths of 0.3 and 0.7 are employed for its low-resolution and high-resolution branches, respectively. Additionally, since during refinement we are aware of the camera pose, thus we also leverage the view-dependent prompting to effectively prevent incorrect face generation when dealing with asymmetric objects. Please noted in the main submission, for certain data we use Control-Net for the refinement, while for the rest we leverage DeepFloyd.


\begin{figure}[!t]
    \begin{center}
    \includegraphics[width=1.0\linewidth]{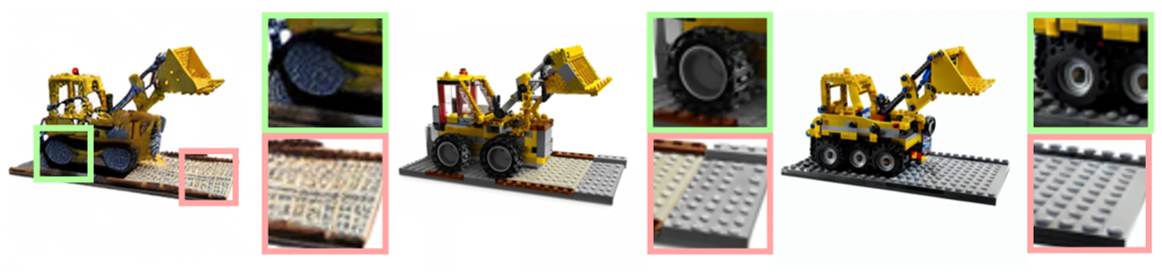}
        \begin{tabularx}{1.0\linewidth}{YYY}
            Zero-1-to-3~\cite{liu2023zero} & DeepFloyd~\cite{deepfloyd} & Control-Net~\cite{Zhang2023ICCV}
        \end{tabularx}
    \end{center}
    \vspace{-1.5em}
    \caption{\textbf{Diffusion refinement to ensure the sampled prior from target distribution is high-quality.} }
    \label{figure:importance_refine}
\end{figure}

We adopt a two-stage approach for training the consistent 3D generator. During the first stage, image-space convolutions are utilized to upscale the raw rendering from $128^2$ triplanes to calculate the adversarial loss. In the second stage, a compact 3D upsampler is employed to lift the triplanes to a $256^2$ resolution. This stage also incorporates patch-level GAN loss and LPIPS for supervision. Notably, we observe a significant quality degradation when training the patch discriminator with adaptive discriminator augmentation (ADA), since the combination of random patch-level augmentation will result in a larger gap compared with original rendering. Hence, we disable ADA in the second stage of training to maintain quality.

To obtain a 3D distribution based on a given reference image and prompt, the entire distillation process takes around three days at a $256^2$ image resolution, which includes 1.5 days for training the 2D upsampler branch, followed by another 1.5 days for finetuning together with the 3D upsampler, using 4 NVIDIA Tesla V100 GPUs. It is worth noting that GAN training can sometimes be unstable, but the integration of our proposed pose pruning strategy could effectively mitigate the need for dense parameter tuning in R1 regularization, for which we consistently set the weight $\gamma=3$ across all experiments.

\begin{figure}[!t]
    \begin{center}
    \includegraphics[width=1.0\linewidth]{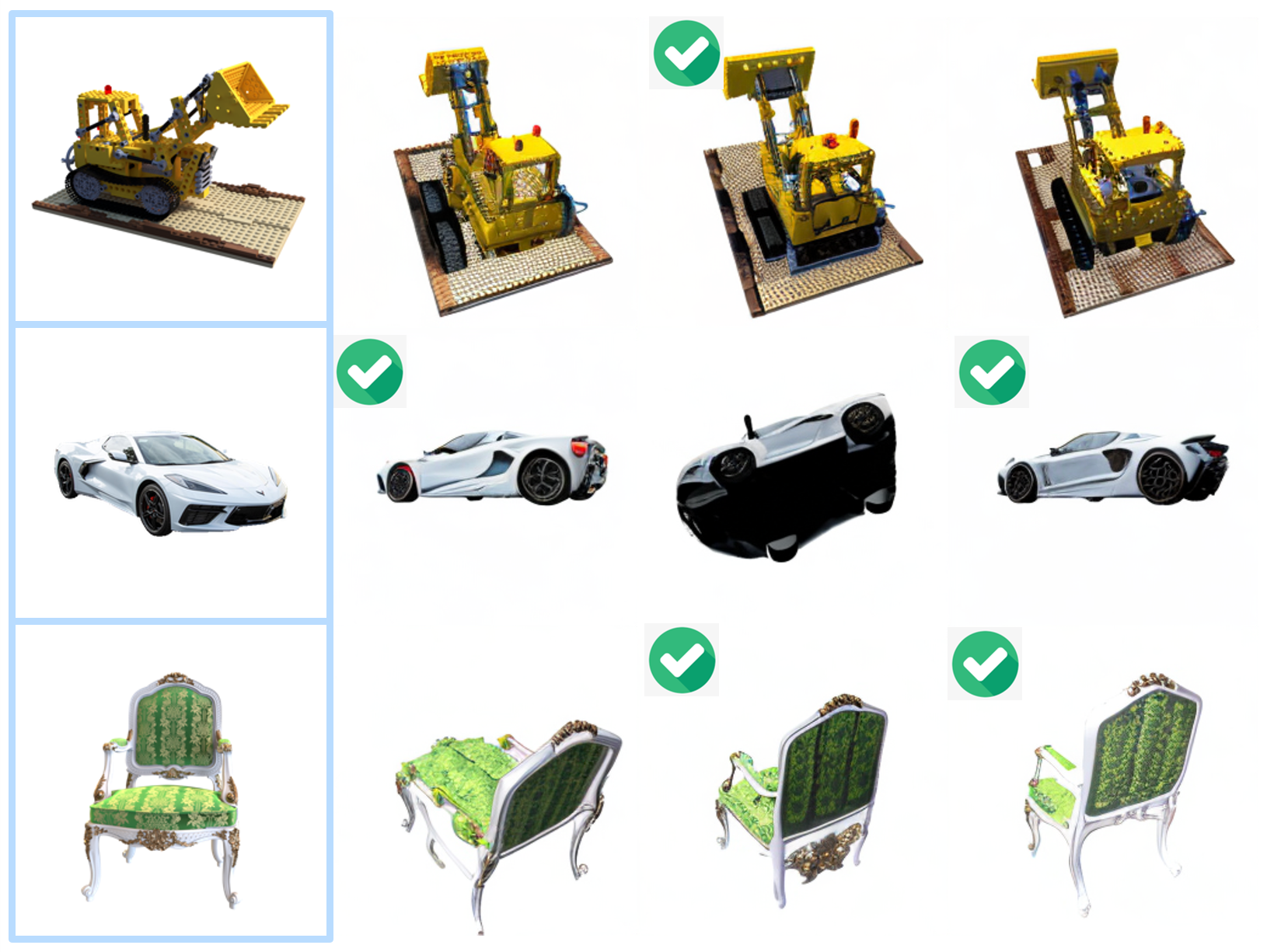}
        \begin{tabularx}{1.0\linewidth}{YYYY}
            Input & Seed 1 & Seed 2 & Seed 3 
        \end{tabularx}
    \end{center}
    \vspace{-1.5em}
    \caption{\textbf{Visualizations of pose pruning.} We show the frequently appeared errors while leveraging the view-dependent diffusion model. By simultaneously considering the parallel sampled results for the same viewpoint, our pruning strategy could effectively filter out the wrong generations.}
    \label{figure:prune_visualize}
\end{figure}

\section{Additional Analysis} \label{Analysis}

\paragraph{Importance of distribution refinement.} The impact of the distribution refinement step on the distillation quality has been demonstrated in the main paper. To further illustrate its importance, we present a visualization in Figure~\ref{figure:importance_refine}. This figure compares directly sampled results from Zero-1-to-3 with those refined by a 2D diffusion model. The refinement process greatly enhances both the quality and diversity of the target distribution, which makes the GAN training process more effective.

\paragraph{Controlling Generation Diversity.} We observed the most significant factor that influences the generation diversity is the noise strength used when sampling the prior. To better demonstrate this observation, we show the refinement results of different diffusion models with continuous noise strengths in Figure.~\ref{figure:noise_strength}. Increasing noise strength results in larger differences from the non-refined input, as it removes more contents from the input. This makes noise strength an effective tool for controlling diversity. However, when using solely text-conditioned diffusion models like DeepFloyd~\cite{deepfloyd}, a higher noise strength (e.g., $s \geq 0.7$) can also induce pose changes. In contrast, the depth-conditioned Control-Net can effectively preserve pose-image correlation due to the geometric constraints from depth. In our main paper, to demonstrate the generality of our adversarial distillation pipeline, we concurrently try the two diffusion models with fixed noise strength. Consequently, the renderings of some models closely resemble the reference image, while others show larger differences yet retain similar semantics.

\begin{figure}[!t]
    \begin{center}
    \includegraphics[width=1.0\linewidth]{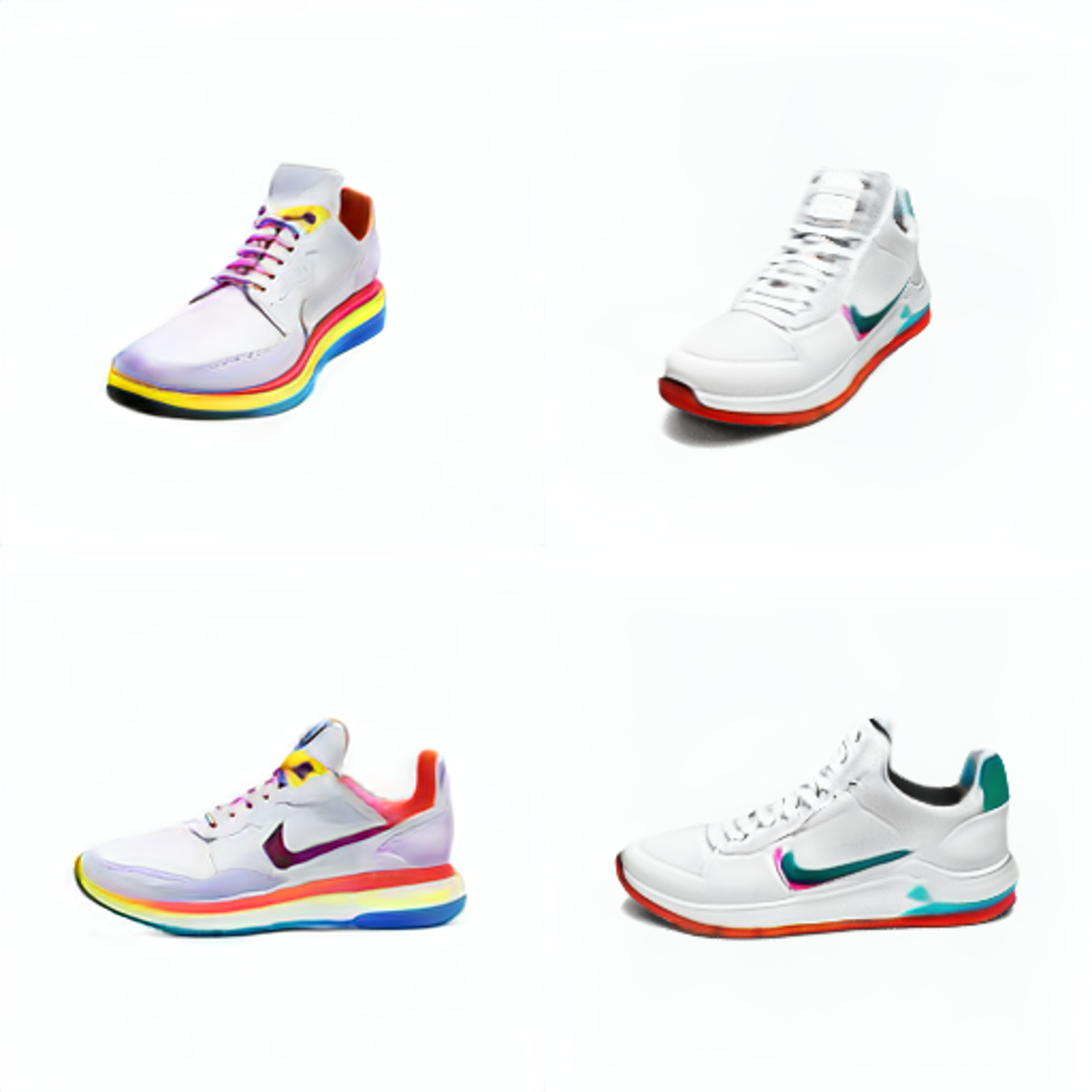}
        \begin{tabularx}{1.0\linewidth}{YY}
         2D upsampling & 3D upsampling 
        \end{tabularx}
    \end{center}
    \vspace{-1.5em}
    \caption{\textbf{Comparisons between 3D generation without consistency training and using our full model.} By directly upsampling the triplanes, the rendering from our model is naturally consistent.}
    \label{figure:importance_consistency}

\end{figure}

\paragraph{Visualization of pose pruning.}  As shown in Figure.~\ref{figure:prune_visualize}, incorporating view-conditioned diffusion models to address sampling bias can result in errors such as mismatched semantics with input views (first row), incorrect camera poses (second row), and wrong 3D structures (third row), all of which will have negative influences on the adversarial training. By considering two aspects including the geometry consistency and semantic consistency, our pruning strategy could effectively filter out these bad samples to make the 3D GAN training stable again even with limited data.



\paragraph{Importance of consistency training.} Limited by the efficiency of volumetric rendering, most 3D GANs~\cite{chan2022efficient, sun2023next3d} rely on image-space convolution for upsampling raw renderings to higher resolutions. While this approach may not result in very visible inconsistencies for face modeling, it is insufficient for synthesizing general 3D objects with a 360° azimuth angle. In Figure\ref{figure:importance_consistency}, we compare 2D and 3D upsampling methods qualitatively. Although 2D upsampling can ensure high rendering quality for each view, the appearance differences become dramatically high when changing camera poses. Our distillation method, however, bakes the abilities of the 2D branch into the 3D branch, which could effectively render multi-view consistent and photorealistic images.



\end{document}